\documentclass{article}

\usepackage{graphicx}
\usepackage{subcaption}

\usepackage[dblblindworkshop, final]{neurips_2025}
\workshoptitle{Machine Learning and the Physical Sciences}

\usepackage[utf8]{inputenc} 
\usepackage[T1]{fontenc}    
\usepackage{hyperref}       
\usepackage{url}            
\usepackage{booktabs}       
\usepackage{amsfonts}       
\usepackage{nicefrac}       
\usepackage{microtype}      
\usepackage{xcolor}         

\newcommand{\resnett}{\texttt{ResNet}$_{train}$}
\newcommand{\resnetr}{\texttt{ResNet}$_{rand}$}

\title{Sparse Methods for Vector Embeddings of TPC Data}

\author{%
Tyler Wheeler$^{1}$ \quad Michelle P.~Kuchera$^{2}$ \quad Raghuram Ramanujan$^{2}$  \\
\textbf{Ryan Krupp}$^1$ \quad \textbf{Chris Wrede}$^1$ \quad \textbf{Saiprasad Ravishankar}$^1$ \\
\textbf{Connor L.~Cross}$^2$ \quad \textbf{Hoi Yan Ian Heung}$^2$ \quad \textbf{Andrew J.~Jones}$^2$ \quad \textbf{Benjamin Votaw}$^2$\\
$^1$ Michigan State University \quad $^2$ Davidson College \\
\texttt{wheele56@msu.edu} \quad \texttt{\{mikuchera,raramanujan\}@davidson.edu}\\
\texttt{\{krupprya,wrede,ravisha3\}@msu.edu}\\
\texttt{\{cocross,iaheung,anjones1,bevotaw\}@davidson.edu}\\
}

\begin{document}

\maketitle

\begin{abstract}
Time Projection Chambers (TPCs) are versatile detectors that reconstruct charged-particle tracks in an ionizing medium, enabling sensitive measurements across a wide range of nuclear physics experiments. We explore sparse convolutional networks for representation learning on TPC data, finding that a sparse ResNet architecture, even with randomly set weights, provides useful structured vector embeddings of events. Pre-training this architecture on a simple physics-motivated binary classification task further improves the embedding quality. Using data from the GAseous Detector with GErmanium Tagging (GADGET) II TPC, a detector optimized for measuring low-energy $\beta$-delayed particle decays, we represent raw pad-level signals as sparse tensors, train Minkowski Engine ResNet models, and probe the resulting event-level embeddings which reveal rich event structure.
As a cross-detector test, we embed data from the Active-Target TPC (AT-TPC)---a detector designed for nuclear reaction studies in inverse kinematics---using the same encoder. We find that even an untrained sparse ResNet model provides useful embeddings of AT-TPC data, and we observe improvements when the model is trained on GADGET data. Together, these results highlight the potential of sparse convolutional techniques as a general tool for representation learning in diverse TPC experiments.
\end{abstract}

\section{Introduction}

Time Projection Chambers (TPCs) are widely used in nuclear physics experiments for their ability to reconstruct three-dimensional charged-particle tracks. By recording ionization signals in a detection medium (typically a gas or liquid), TPCs enable sensitive studies of rare decays and reaction dynamics across a broad range of energies and beam conditions \citep{Hilke_2010}. However, these benefits come with substantial computational challenges: TPCs produce massive, high-dimensional datasets that are inherently sparse and vary significantly across detector geometries and experimental goals.

Traditional approaches to analyzing TPC data often rely on detector-specific pipelines, limiting the development of generalizable tools that can be applied across experiments. Deep learning approaches have been successful for TPC data, with deep convolutional networks being a common choice for TPC ``images'' \citep{KUCHERA2019156,  SOLLI2021165461, PhysRevD.103.092003, Carloni_2022, WHEELER2025170659, DEY2025170002}, but have thus far required independent training for each task of interest. They also suffer from computational inefficiency when applied to sparse high-dimensional inputs.

\begin{figure}[htbp] 
    \centering 
    \begin{subfigure}[b]{0.4\textwidth}
        \centering
        \includegraphics[width=\textwidth]{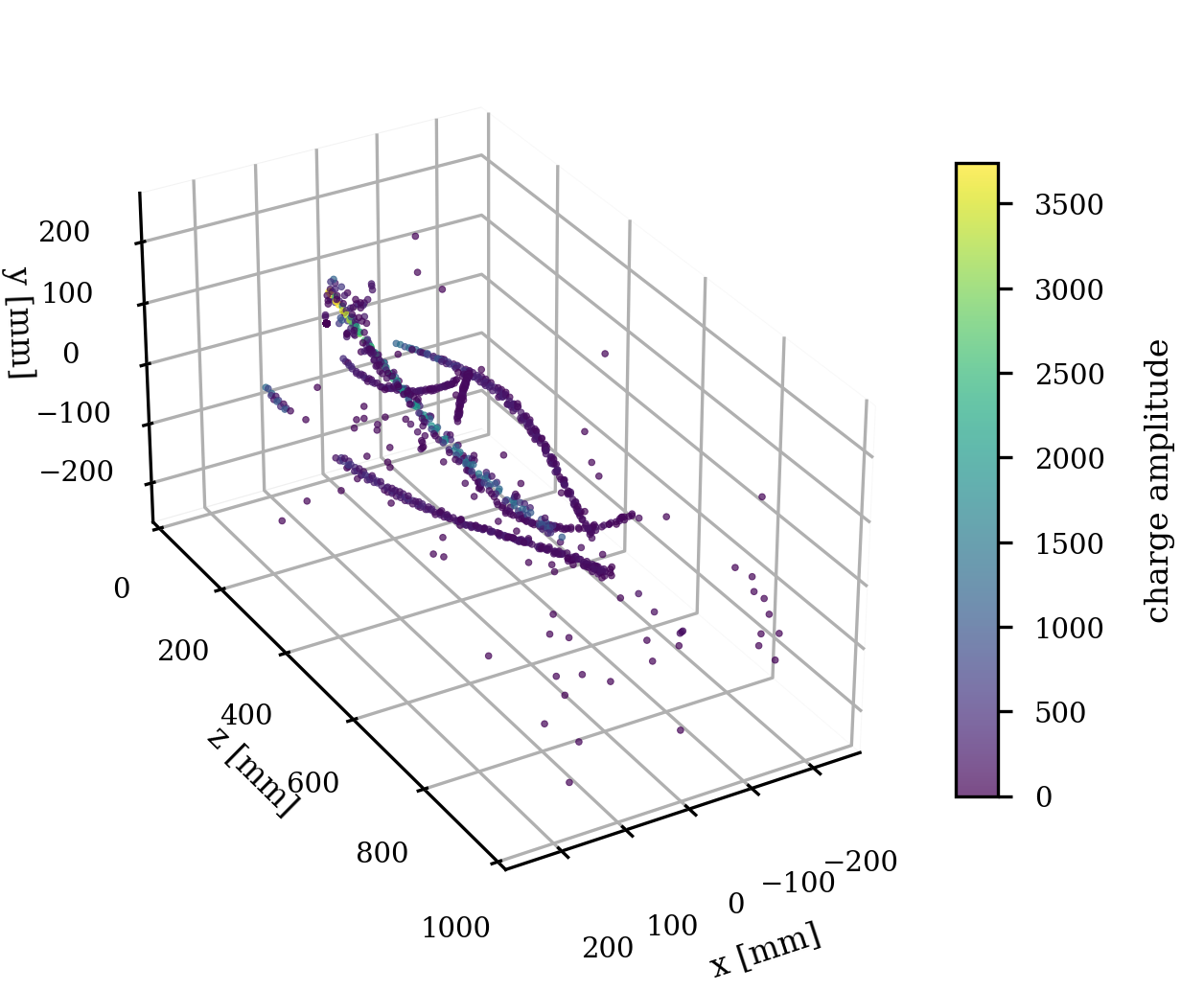}
        \label{fig:attpc}
    \end{subfigure}
    \qquad \qquad
    \begin{subfigure}[b]{0.49\textwidth}
        \centering
        \includegraphics[width=\textwidth]{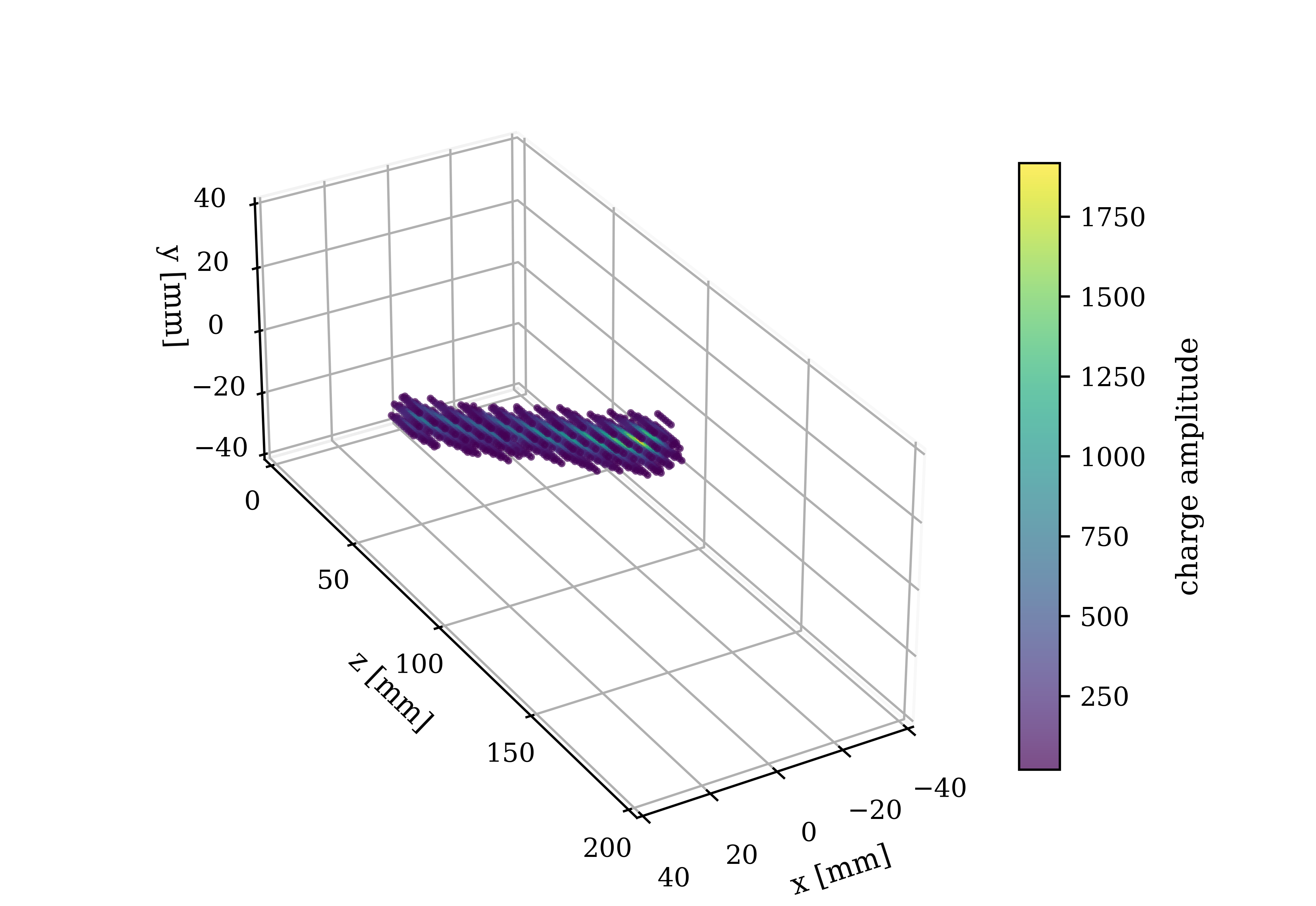} 
        \label{fig:second_plot}
    \end{subfigure}
    \caption{Left: an example of a 4-track AT-TPC event. Right: an example of a 1600 keV proton event from the GADGET II TPC.}
    \label{fig:side_by_side}
\end{figure}

In this work, we explore the use of sparse convolutional neural networks, implemented via the Minkowski Engine framework \citep{Choy_2019_CVPR}, to process raw TPC data efficiently and extract meaningful event representations. We are particularly interested in whether such models can transfer across different TPC systems---rather than purely optimizing classifier performance---thereby motivating the eventual development of general-purpose TPC foundation models. We focus primarily on the GAseous Detector with GErmanium Tagging (GADGET~II) TPC \citep{PhysRevC.110.035807}, detailing how raw pad-level signals are represented as sparse tensors, trained with ResNet models, and analyzed in a latent space. To demonstrate the richness of the embedding space, we also embed data from the $^{16}$O $+ \alpha$ experiment that used the Active-Target TPC (AT-TPC) \citep{AYYAD2020161341}. The two detectors differ markedly in geometry and scientific goals—GADGET~II is compact and optimized for low-energy $\beta$-delayed charged particle decays, whereas the AT-TPC provides wide solid-angle coverage for inverse-kinematics reactions and is surrounded by a solenoid magnet that provides up to 2 T of magnetic field—allowing us to assess how well sparse tensor methods carry across TPC designs. Example data from both detectors are shown in Figure~\ref{fig:side_by_side}. We would like to draw the reader's attention to the visual contrast between these event topologies: tracks from GADGET II and the AT-TPC look like data from entirely unrelated domains. It is thus remarkable that transfer learning between such distinct detectors is effective. A concise summary of our findings is provided in the Results and Conclusion sections. 

\section{Methods}
\subsection{Network Architecture}

We employ sparse convolutional neural networks implemented with the Minkowski Engine to process four-dimensional point cloud data from TPCs, where each point consists of spatial coordinates and an associated charge $(x, y, z, q)$. Convolutions operate only in the three spatial dimensions, while $q$ is treated as a feature channel. This preserves pad-level sparsity and avoids the cost of dense voxelization.

The backbone is a shallow residual network (ResNet14) \citep{he2015deepresiduallearningimage} adapted for sparse inputs. Its main components are summarized in Table~\ref{tab:architecture}. The design includes an initial stem, four residual stages, and a final pre-pooling block followed by global max pooling and a fully connected head. This configuration uses ten convolutional layers and one linear layer. We found that deeper variants (e.g., ResNet50 with bottleneck blocks) increased training time without clear performance gains

A key advantage of this architecture is its ability to handle variable-length inputs natively. Unlike many implementations (e.g., typical PointNet \citep{pointnet} training pipelines) that require padding or truncation, sparse ResNets flexibly accommodate events with widely varying numbers of active channels. This is particularly important for TPC data, where track lengths and hit multiplicities can vary substantially both within and across detectors.

\begin{table}[t]
\centering
\caption{Summary of the sparse ResNet14 architecture. Convolutions are sparse $3{\times}3{\times}3$ operations unless otherwise noted.}
\label{tab:architecture}
\setlength{\tabcolsep}{8pt}
\begin{tabular}{@{}lll@{}}
\toprule
\textbf{Stage} & \textbf{Description} & \textbf{Key Operations} \\
\midrule
Stem & Input processing & Sparse conv (stride 2), BN, ReLU, max pool ($2^3$) \\
Residual Stage 1--4 & Feature extraction & $2\times$ BasicBlock per stage (ReLU); stride 2 at start \\
Downsampling & Dimension alignment & $1^3$ sparse conv when required \\
Pre-pooling block & Feature refinement & Dropout ($0.8$), sparse conv (stride 3), BN, GELU \\
Global pooling & Feature aggregation & Global max pool \\
Head & Output & Fully connected linear layer \\
\bottomrule
\end{tabular}
\end{table}

\begin{table}[t]
\centering
\caption{Key training and optimization settings.}
\label{tab:train}
\setlength{\tabcolsep}{6pt}
\begin{tabular}{@{}ll@{}}
\toprule
\textbf{Item} & \textbf{Setting} \\
\midrule
Loss & Cross-entropy \\
Optimizer & Adam (LR $5\times 10^{-4}$, weight decay $10^{-4}$) \\
Scheduler & Cosine annealing ($T_{\max}=13$) \\
Batching & Variable-length events; batch size $64$ \\
Regularization & Gradient clipping (max-norm $=1.0$), dropout $p=0.8$ \\
Epochs & $15$ (best model by validation loss) \\
Class imbalance & Weighted random sampler \\
Metrics & Accuracy, precision, recall, macro-F1 \\
\bottomrule
\end{tabular}
\end{table}

\subsection{Training Setup}

\begin{figure}[t]
  \centering
  \includegraphics[width=\linewidth]{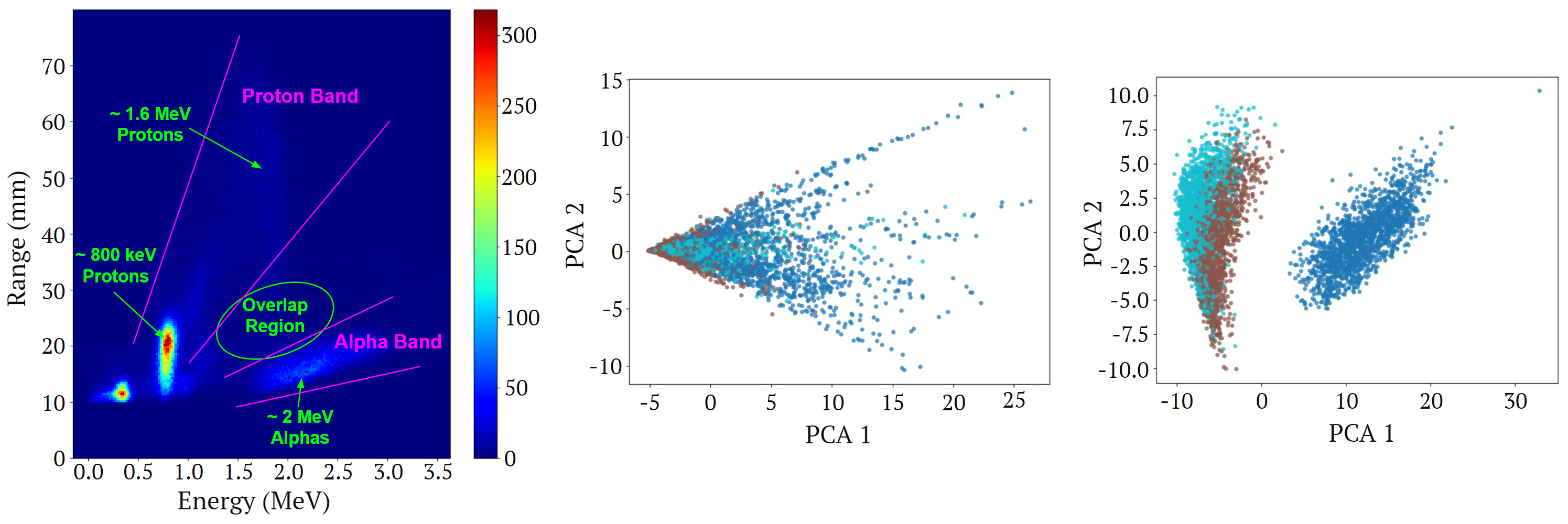}
  \caption{Left: Range vs energy plot for FRIB experiment 21072 (1 hr run). Distinct proton and alpha bands are visible; gating cuts on these bands provided training labels for models, while events in the overlap region were excluded. Middle: PCA of embeddings from \resnetr~. Right: PCA of embeddings from \resnett~. In PCA plots: 800 keV $p$, 1600 keV $p$, and 2 MeV $\alpha$; brown, light blue, and dark blue, respectively.}
  \label{fig:r_v_e}
\end{figure}

The sparse ResNet14 model is trained with four input channels $(x, y, z, q)$ to perform binary proton–alpha classification on GADGET~II data; we refer to this pretrained model as \resnett. Labels are derived by gating distinct bands in the range–energy plane (Figure~\ref{fig:r_v_e}). Inputs are raw pad-level hits quantized with a voxel size of $0.05$. Batching uses Minkowski Engine’s \texttt{batched\_coordinates} without padding. The best checkpoint is selected using the lowest validation loss, and training converges within $\sim$15 epochs, achieving $\approx 0.995$ accuracy and $\approx 0.99$ F1. Key hyperparameters and optimization settings are summarized in Table~\ref{tab:train}.


\subsection{Latent Embedding and Analysis}

We extract latent vectors from the penultimate layer of our ResNet14 model trained solely on the binary proton--alpha classification task (\resnett). Additionally, we also investigate the embeddings produced by a randomly initialized ResNet14 that undergoes no further training (\resnetr). Prior research with dense CNNs has uncovered that such randomly weighted networks produce surprisingly effective embeddings  \citep{deepimageprior,randomwts,randomwts2} and we use these as a strong baseline for comparison. 

To investigate the information encoded in the latent space of our models, we train a linear probe \citep{probing}---specifically, a linear support vector machine---on various classification tasks that are distinct from the training task. Deliberately choosing a low capacity model as the probe allows us to test whether the embeddings make useful information about events accessible in a straightforward, readily interpretable form.


To visualize the structure of the embedding space, we apply principal component analysis (PCA) \citep{jolliffe2002principal} to reduce the high-dimensional latent vectors to their first two principal components. This linear projection captures the directions of maximum variance, providing a view of the embeddings that is legible to humans and allowing us to directly see whether architectural choices and/or pretraining induce separation among event types. 



\section{Results}


\subsection{Linear Probing}

To probe the information embedded in the latent space of our models, we use a linear classifier to distinguish between categories that were not used in training. We probe embedded representations of data both from GADGET II and from the AT-TPC using both \resnett~and \resnetr. 

We train a linear support vector machine on classification tasks that are distinct from the training task for \resnett. 
For GADGET II data, the linear probe achieved high performance on a three-class separation task---distinguishing between $800\,\mathrm{keV}$ protons, $1600\,\mathrm{keV}$ protons, and $2\,\mathrm{MeV}$ alphas---that is more challenging than the binary classification task used for pretraining. Accuracy and F1 both reached $0.97$ when using the \resnett~embeddings, compared to $0.85$ for \resnetr~(see Table~\ref{tab:model_performance}). Both numbers are significantly higher than the corresponding figures for the na\"ive model that always predicts the mode: accuracy of $0.33$ and an F1 score of $0.17$. This indicates that the embeddings capture physically meaningful substructure beyond the binary training labels, enabling linear separability of multiple decay channels. Moreover, the higher performance of the \resnett~embeddings suggests that there is value in pretraining on in-domain data, even if it's on a different task.

For AT-TPC data, embeddings from \resnetr~and \resnett~were probed using a track counting task  ($\{0,1,2\}$ vs. $\{3\}$ vs. $\{4,5\}$ tracks). 
The results (see Table~\ref{tab:model_performance}) show that latent features learned on one detector retain their utility when transferred to a different system with vastly different geometry and physics goals.

It is important to note that our objective in these experiments is not to realize the highest-performing classifier. With further optimization and engineering effort, higher classification accuracy could undoubtedly be achieved. Rather, our intent is to demonstrate that this form of transfer learning is viable in practice and that it provides a basis for developing broadly useful TPC foundation models.

\subsection{PCA Analysis}
PCA projections of the GADGET II embeddings are shown in Figure~\ref{fig:r_v_e}. Even without training (\resnetr), the architecture produces embeddings with some degree of ordering: classes are partially separate in the 2D projection, reflecting the architectural bias of sparse convolutions toward learning from local spatial continuity. 

Embeddings from \resnett, which was optimized only for binary proton–alpha discrimination, exhibit an even cleaner separation of all three particle classes (800 keV $p$, 1600 keV $p$, 2 MeV $\alpha$). This shows that pretraining on a simple physics-driven task yields representations with richer event structure than the labels alone, producing a latent space where multi-class tasks can become linearly separable. 

\begin{figure}
    \centering
    \includegraphics[width=\linewidth]{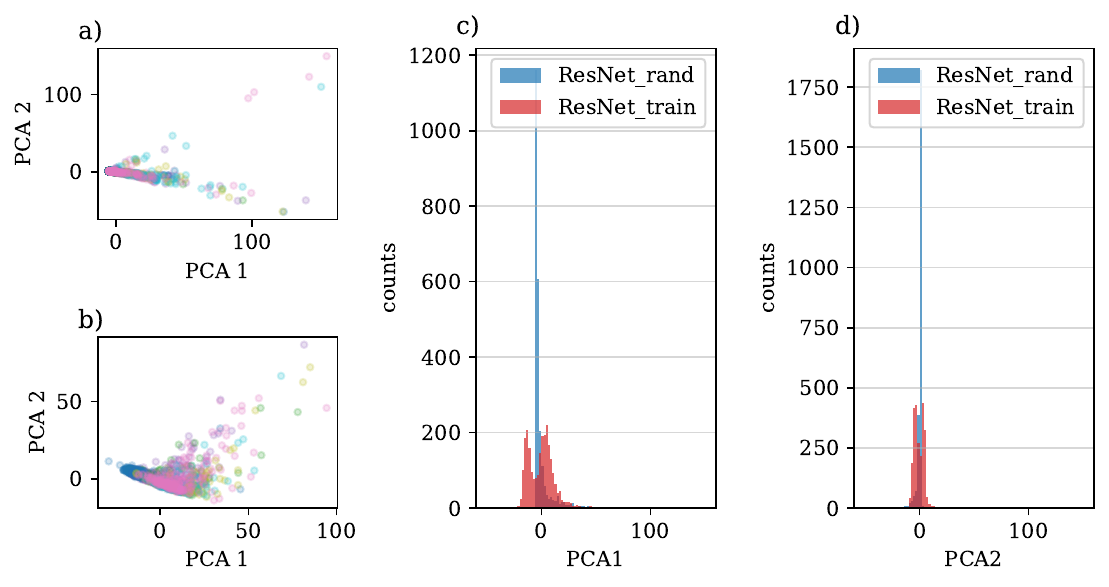}
    \caption{Visualization of AT-TPC ResNet embeddings. Plot a) shows the first two principal components (PCA1 and PCA2 respectively) of embeddings from \resnetr. Plot b) depicts PCA2 vs.~PCA1 from \resnett. Plots c) and d) present the distribution of latent embeddings for PCA1 and PCA2, respectively. }
    \label{fig:attpc_PCA}
\end{figure}

We further applied PCA to embeddings of AT-TPC data (Figure~\ref{fig:attpc_PCA}). For \resnetr, projections again show weak but nontrivial organization, with narrow distributions and limited separation. In contrast, embeddings obtained by running AT-TPC events through the GADGET-trained encoder (\resnett) show more dispersed embeddings in PCA space, allowing for more successful separation. This demonstrates that features learned from GADGET II not only enhance latent separation in-domain but also transfer effectively across detectors.

\begin{table}[h!]
\centering
\resizebox{0.9\textwidth}{!}{%
\begin{tabular}{|l|ccc|ccc|}
\hline
\textbf{Domain} & \multicolumn{3}{c|}{\textbf{Accuracy}}  & \multicolumn{3}{c|}{\textbf{F1 Score}} \\ \hline
 &  \resnett & \resnetr & Na\"ive & \resnett & \resnetr & Na\"ive \\ \hline
GADGET II & \textbf{0.97} & 0.85  & 0.33 & \textbf{0.97} & 0.85 & 0.17 \\ \hline
AT-TPC   & \textbf{0.74} & 0.58 & 0.48  & \textbf{0.70} & 0.53 & 0.31 \\ \hline
\end{tabular}
} 
\vspace{5pt}
\caption{Linear probe results on GADGET II three-class particle classification and AT-TPC track counting tasks. The untrained architecture (\resnetr) provides non-trivial structure; pretraining on GADGET II data (\resnett) improves both in-domain and out-of-domain transfer.}
\label{tab:model_performance}
\end{table}

\section{Conclusion}

We showed that embeddings learned by sparse convolutional networks trained on GADGET II data not only enable accurate classification of decay events within that detector, but also transfer effectively to new tasks, and even to different TPCs with distinct geometries and physics goals (AT-TPC). While even an untrained architecture yields embeddings with useful structure, pretraining on a straightforward GADGET II proton-alpha classification task substantially improves the utility of the embeddings. These results highlight the value of sparse tensor methods for analyzing and processing TPC data and their promise for building cross-domain foundation models for TPCs.

\section{Acknowledgments}
This work was supported by the U.S. Department of Energy, Office of Science, under award No. DE-SC0024587 and the National Science Foundation Office of Advanced Cyberinfrastructure award No. OAC-2311263.

\setlength{\bibhang}{0pt}
\bibliographystyle{plainnat} 
\bibliography{r} 

\end{document}